\documentclass[10pt,twocolumn,letterpaper]{article}

\usepackage[pagenumbers]{wacv} 

\usepackage{graphicx}
\usepackage{amsmath}
\usepackage{amssymb}
\usepackage{booktabs}
\usepackage{mathtools}
\usepackage{array}
\usepackage[pagebackref,breaklinks,colorlinks]{hyperref}

\usepackage[capitalize]{cleveref}
\crefname{section}{Sec.}{Secs.}
\Crefname{section}{Section}{Sections}
\Crefname{table}{Table}{Tables}
\crefname{table}{Table}{Tables}

\def\wacvPaperID{1212} 
\def\confName{WACV}
\def\confYear{2025}

\def\softmax{\mathop{\mathrm{Softmax}}\nolimits}

\def\comment[#1:#2]{\textbf{[#1: #2]}}

\begin{document}

\title{Harmonizing Attention: Training-free Texture-aware Geometry Transfer}

\author{
    \begin{tabular*}{0.8\textwidth}{@{\extracolsep{\fill}}c c c}
        {Eito Ikuta}$^{1}$ & {Yohan Lee}$^{2}$ & {Akihiro Iohara}$^{1}$
        \\ \small{eito.ikuta@datagrid.co.jp} & \small{lee.yohan.83w@st.kyoto-u.ac.jp} & \small{akihiro.iohara@datagrid.co.jp}
    \end{tabular*}\\[1.5em]
    \begin{tabular*}{0.5\textwidth}{@{\extracolsep{\fill}}c c}
        {Yu Saito}$^{1}$ & {Toshiyuki Tanaka}$^{2}$
        \\ \small{yu.saito@datagrid.co.jp} & \small{tt@i.kyoto-u.ac.jp}
    \end{tabular*}
    \\[2em] 
    \begin{tabular*}{0.5\textwidth}{@{\extracolsep{\fill}}c c}
        \textsuperscript{1}{DATAGRID Inc.} & \textsuperscript{2}{Kyoto University}
    \end{tabular*}
}



\twocolumn[{%
\renewcommand\twocolumn[1][]{#1}%
\maketitle
\begin{center}
    \centering
    \captionsetup{type=figure}
    \includegraphics[width=\textwidth]{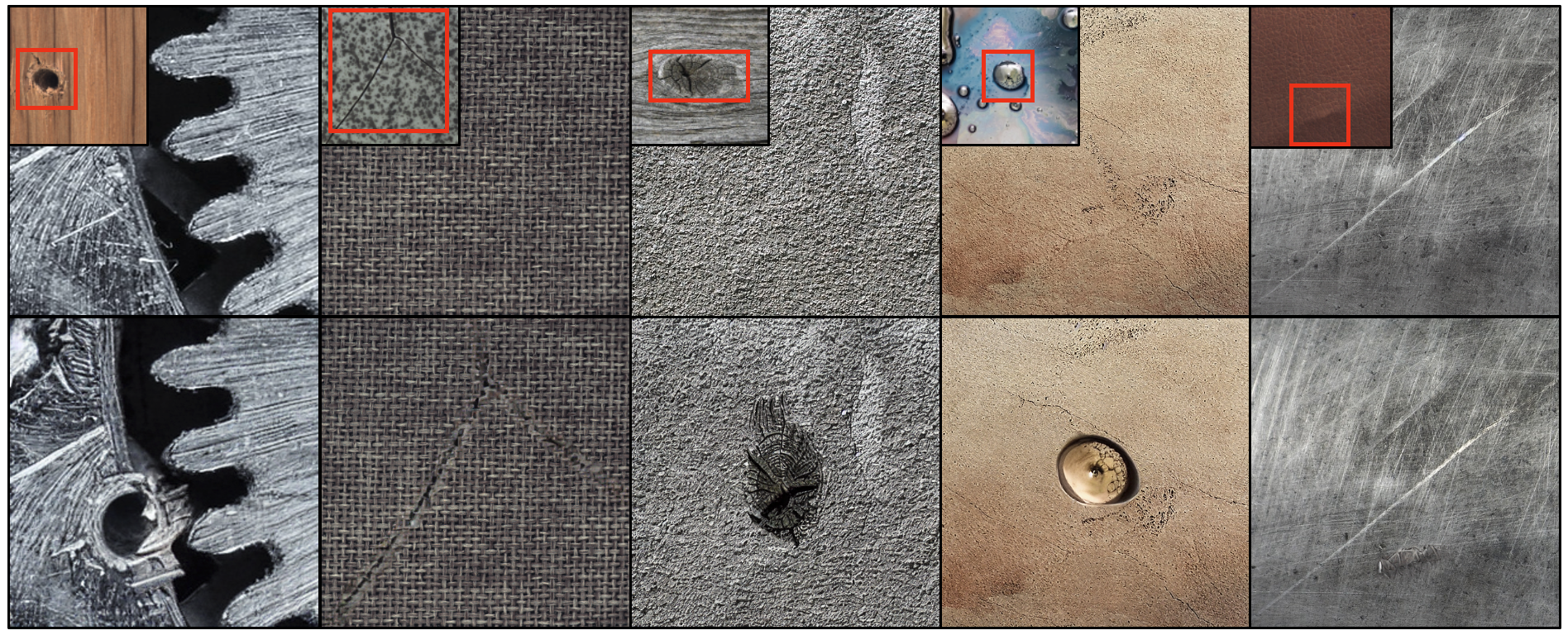}
    \captionof{figure}{Examples of texture-aware geometry transfer. Our method seamlessly transfers geometric features (holes, cracks, droplets, etc.) from source images (red rectangles in the insets of the top row) to target images with different surface textures (main images in the top row). }
    \label{fig:eyecatch}
\end{center}%
}]



\begin{abstract}
Extracting geometry features from photographic images independently of surface texture and transferring them onto different materials remains a complex challenge. In this study, we introduce Harmonizing Attention, a novel training-free approach that leverages diffusion models for texture-aware geometry transfer. Our method employs a simple yet effective modification of self-attention layers, allowing the model to query information from multiple reference images within these layers. This mechanism is seamlessly integrated into the inversion process as Texture-aligning Attention and into the generation process as Geometry-aligning Attention. This dual-attention approach ensures the effective capture and transfer of material-independent geometry features while maintaining material-specific textural continuity, all without the need for model fine-tuning.
\end{abstract}

\section{Introduction}
\label{sec:intro}
In computer vision, image harmonization is the task of seamlessly integrating a foreground object from one image into the background of another to produce a cohesive composite. A significant challenge is achieving visual harmony by adjusting the foreground's appearance to align with the background. Traditional methods~\cite{Cohen-Or2006, dragdroppaste, lalonde2007using, Perez2003, Pitie2005, Reinhard2001, Sunkavalli2010, error-tolerant, Xue2012, Zhu2015} have primarily addressed image harmonization by focusing on color and illumination adjustments.

Recent advances in deep learning and diffusion models~\cite{Cong2020, Cun2020, Guo_2021_ICCV, Jiang_2021_ICCV, Ling2021, cong2021bargainnet, sofiiuk2021foreground, guo2022transformer, Hang2022SCSCo, Song_2023_CVPR, TF-ICON} have enabled sophisticated image composition and harmonization, as exemplified by ObjectStitch~\cite{Song_2023_CVPR} and TF-ICON~\cite{TF-ICON}. These methods primarily focus on style transfer, encompassing a broad range of attributes such as color schemes, brushstrokes, textures, and patterns that collectively determine an image's visual appearance.
However, the selective transfer of geometrical features—which we term ``geometry" in this paper—such as holes, cracks, droplets, and dents from one material to another, independently of material-specific surface texture (e.g., wood grain or fabric weave), remains a complex challenge. To the best of our knowledge, existing techniques have not successfully addressed this issue. Such geometry transfer is crucial for creating realistic and aesthetically pleasing composite images, particularly in applications requiring the integration of complex textures from different materials, like transferring wood hole geometry onto metal surfaces. Traditional and recent deep-learning-based image harmonization methods struggle with geometry transfer due to their focus on color, illumination, and style adjustments of the object itself. This limitation highlights the need for a fundamentally different approach that can effectively decouple geometry from material-specific textures in image harmonization tasks.

To address this challenge, we propose Harmonizing Attention, a novel training-free diffusion-model-based approach for texture-aware geometry transfer. The key aspect of this method is a custom attention mechanism comprising Texture-aligning Attention and Geometry-preserving Attention, enabling reference to both source geometry and target texture information. By replacing the self-attention layers in the diffusion model with these custom attention layers during inversion and generation, we successfully incorporate geometry and texture from different images, as demonstrated in ~\cref{fig:eyecatch}. Notably, since the Harmonizing Attention framework leverages pretrained diffusion models to perform geometry transfer, it requires neither model tuning nor prompt exploration for each dataset.

Our contributions are summarized as follows:
\begin{enumerate}
\item We introduce texture-aware geometry transfer, a novel approach for integrating geometry details from different materials into cohesive composite images.
\item We develop Harmonizing Attention, a training-free mechanism enabling models to query information from multiple reference images within self-attention blocks, effectively capturing and transferring material-independent geometry while preserving material-specific textures.
\item Our experiments demonstrate that Harmonizing Attention produces more harmonious and realistic composite images compared with existing techniques.
\end{enumerate}

\section{Related Work}
\label{sec:related_work}

\paragraph{Image Harmonization.}
Conventionally, image harmonization ~\cite{Cohen-Or2006, dragdroppaste, lalonde2007using, Perez2003, Pitie2005, Reinhard2001, Sunkavalli2010, error-tolerant, Xue2012, Zhu2015} has focused on color-to-color transformations to match visual appearances. These methods can be further divided into non-linear transformations~\cite{Sunkavalli2010, Xue2012} and linear transformations~\cite{lalonde2007using, Reinhard2001, Zhu2015}. Recently, deep-learning-enabled image harmonization emerged~\cite{Cong2020, Cun2020, Guo_2021_ICCV, Jiang_2021_ICCV, Ling2021, cong2021bargainnet, sofiiuk2021foreground, guo2022transformer, Hang2022SCSCo} and more recently diffusion-model-based harmonization techniques~\cite{Song_2023_CVPR,TF-ICON} have been developed. One notable example is TF-ICON~\cite{TF-ICON}, which equips attention-based text-to-image diffusion models for image harmonization, enabling cross-domain image-guided composition.

\paragraph{Paintely Image Harmonization.}
In the context of image harmonization, a similar task called painterly image harmonization~\cite{DIH,4409107, GP_GAN,DIM_dual,hierachical_harmonization, composite_photograph} has been studied, and it might be another promising candidate technique for geometry transfer. The painterly image harmonization technique integrates a photographic foreground into an artistic background, resulting in a visually coherent painting. PHDiffusion~\cite{PHDiffusion} represents a significant pivot by adapting the GAN framework (PHDNet~\cite{DIH}) to a diffusion model with an adaptive encoder. Furthermore, TF-GPH~\cite{TF-GPH}, which utilizes an image-wise attention-sharing mechanism for general painterly harmonization, is proposed for training-free general painterly harmonization and provides flexible options for attention-based image editing methods.


\section{Methods}
\label{sec:methods}
\begin{figure*}[ht]
\centering
\includegraphics[width=0.9\linewidth]{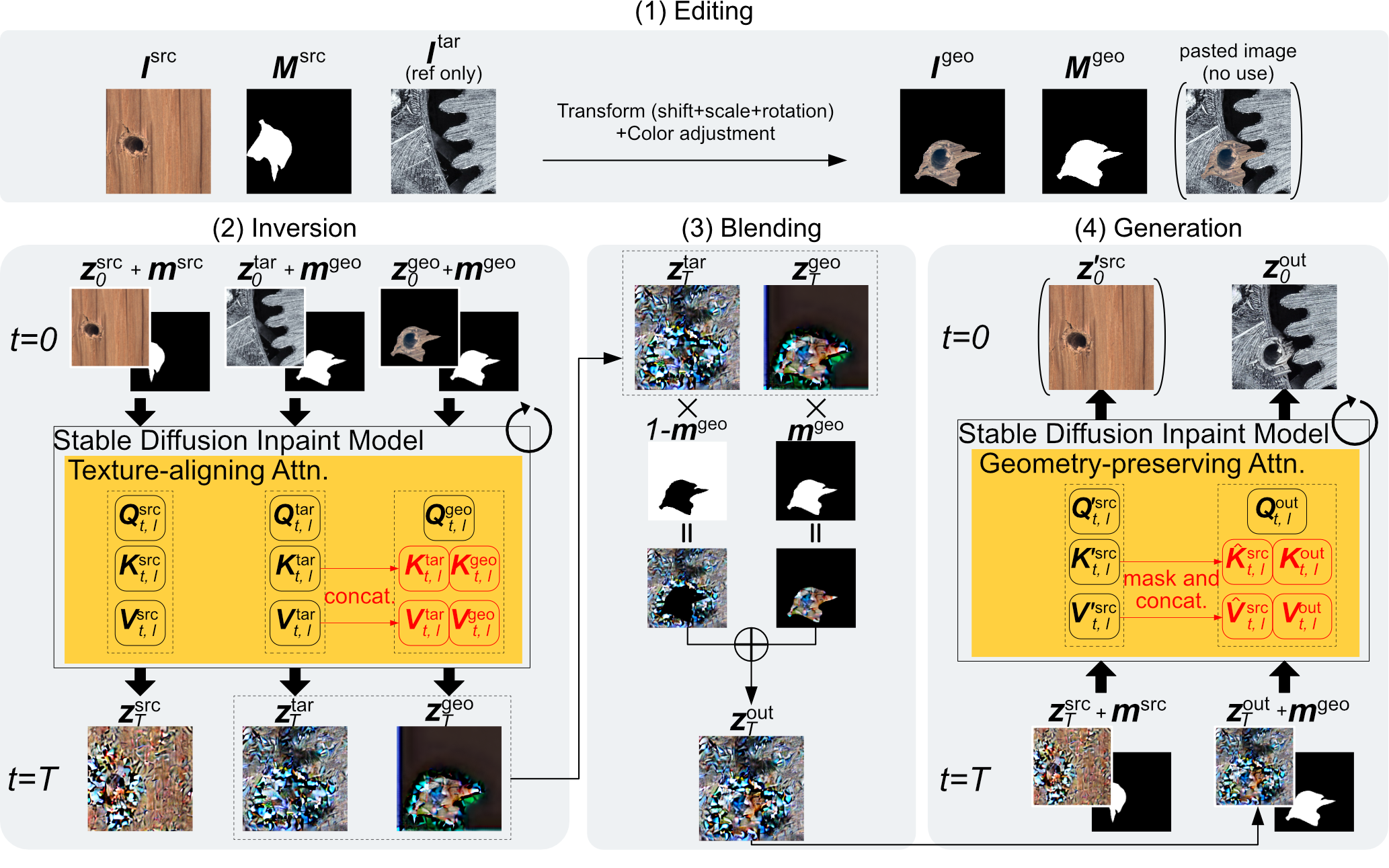}
\caption{
    The overview of Harmonizing Attention framework.
    The process consists of four main stages: 
    (1) Editing: The masked region of the source image $\boldsymbol{M}^\mathrm{src} \odot \boldsymbol{I}^\mathrm{src}$ is transformed, positioned, and color-adjusted to create a geometry image $\boldsymbol{I}^\mathrm{geo}$ and a corresponding mask $\boldsymbol{M}^\mathrm{geo}$.
    (2) Inversion: The process commences with the compression of the source, target, and geometry images via VAE, yielding clear latents $\boldsymbol{z}^\mathrm{src}_0$, $\boldsymbol{z}^\mathrm{tar}_0$, and $\boldsymbol{z}^\mathrm{geo}_0$ respectively.
    Subsequently, these latents, along with the source and geometry masks resized to the latent space, are fed into an SD inpainting model.
    This step results in the generation of inverted noisy latents $\boldsymbol{z}^\mathrm{src}_T$, $\boldsymbol{z}^\mathrm{tar}_T$, and $\boldsymbol{z}^\mathrm{geo}_T$.
    Here we replace the self-attention computation for the geometry image with Texture-aligning Attention, which incorporates information from the target image to align the geometry image with the target domain. 
    (3) Blending: The latents of the target and geometry images are combined using the geometry mask. 
    (4) Generation: The blended latents undergo denoising, with modified self-attention named Geometry-preserving Attention, to reference the source image, preserving the geometry while ensuring seamless integration.
    For the sake of enhanced clarity and readability, all latents are presented in pixel space, rather than in the VAE latent space.}
\label{fig:overview}
\vspace{-0.4cm} 
\end{figure*}

Our objective is to synthesize an image $\boldsymbol{I}^\mathrm{out}$ that seamlessly integrates surface geometry information from a source image $\boldsymbol{I}^\mathrm{src}$ into a target background image $\boldsymbol{I}^\mathrm{tar}$.
This integration is guided by a 0-1-valued foreground mask $\boldsymbol{M}^\mathrm{src}$, preserving the geometry of $\boldsymbol{M}^\mathrm{src} \odot \boldsymbol{I}^\mathrm{src}$ while maintaining textural continuity with $\boldsymbol{I}^\mathrm{tar}$, independently of the source's original texture, 
where $\odot$ denotes the Hadamard (elementwise) product. 
To achieve this, we propose a novel framework called Harmonizing Attention.
As illustrated in ~\cref{fig:overview}, this framework leverages Stable Diffusion (SD)~\cite{StableDiffusion} and encompasses both inversion and generation processes.
The key aspect of our approach is modifying self-attention computation during both inversion and generation to query additional information from $\boldsymbol{I}^\mathrm{src}$ and $\boldsymbol{I}^\mathrm{tar}$, enabling a more coherent and context-aware transfer process.
Furthermore, to enhance textural continuity within the target region, we utilize the SD inpainting model rather than the text-to-image model.

\paragraph{Editing.}
\label{par:Editing}
The initial stage of our framework involves preparation of a geometry image $\boldsymbol{I}^\mathrm{geo}$, which serves as a cornerstone for the subsequent image generation process.
A geometry image is simply an image patch 
of a source image cropped by a 0-1 source mask $\boldsymbol{M}^\mathrm{src}$, with affine transformations $\mathcal{T}$ 
(shifting, scaling, and rotation) according to the user's needs, 
as well as the color adjustment described below 
(see the uppermost panel in ~\cref{fig:overview}). 
The ultimate objective of our framework is to seamlessly synthesize this geometry image with the target image.

The color adjustment of the geometry image plays a crucial role in that it facilitates the inversion process to work properly.
By aligning the color profile of the source region with that of the target image, we expect that it will be easier to bring the geometry image closer to the visual domain of the target through the inversion procedure.
Because the geometry information is expressed through localized color differences rather than absolute values, we employ a simple uniform color shift, although more sophisticated color adjustment methods would also be feasible.
Specifically, we compute this shift by calculating the difference between two color averages: (1) $\boldsymbol{c}^\mathrm{src} \in \mathbb{R}^3$, which is the mean color of the source image $\boldsymbol{I}^\mathrm{src}$ within the region formed by subtracting the source mask $\boldsymbol{M}^\mathrm{src}$ from its dilated version (i.e., the boundary of the complementary mask $\boldsymbol{M}^\mathrm{src}_{\mathrm{comp}}$ in mathematical morphology, 
where the complementary mask $\boldsymbol{M}_{\mathrm{comp}}$ 
of a mask $\boldsymbol{M}$ is defined by replacing 0 with 1 and vice versa in $\boldsymbol{M}$);
(2) $\boldsymbol{c}^\mathrm{tar} \in \mathbb{R}^3$, which is the mean color of the target image $\boldsymbol{I}^\mathrm{tar}$ within an analogous region, defined by subtracting the transplantation-area mask  $\boldsymbol{M}^\mathrm{geo} \coloneq \mathcal{T}(\boldsymbol{M}^\mathrm{src})$  from its dilated version, where the transplantation-area mask $\boldsymbol{M}^\mathrm{geo}$ defines the transplantation region in the target-image coordinate.
This process is represented by the following equation:
\begin{equation}
  \boldsymbol{p}^\mathrm{geo}_{x,y} = \boldsymbol{\tilde{p}}^\mathrm{geo}_{x,y} + aM^\mathrm{geo}_{x,y}(\boldsymbol{c}^\mathrm{tar} - \boldsymbol{c}^\mathrm{src}),
  \label{eq:color-adjustment-in-editing}
\end{equation}
where $\boldsymbol{\tilde{p}}^\mathrm{geo}_{x,y}, \boldsymbol{p}^\mathrm{geo}_{x,y} \in \mathbb{R}^3$ represent the pixel values at coordinates $(x, y)$ in the image before and after color adjustment, respectively.
$M^\mathrm{geo}_{x, y} \in \{0, 1\}$, value of $\boldsymbol{M}^\mathrm{geo}$ at coordinates $(x, y)$, becomes 1 only if the pixel $(x,y)$ 
in the target image is within the transplantation region.
Equation~\eqref{eq:color-adjustment-in-editing} then shows 
that the color adjustment is done only within 
the transplantation region. 
The scalar parameter $a \in \lbrack0, 1\rbrack$ controls the strength of the color adjustment.
It is anticipated that employing a small value of $a$ would result in inadequate integration of the transplantation region with the target image, while a large value of $a$ is expected to compromise the integrity of the source geometry.

\paragraph{Inversion.}
The geometry image $\boldsymbol{I}^\mathrm{geo}$ inherently retains substantial textural information from the source image.
Consequently, a direct pasting of the geometry image onto the target image $\boldsymbol{I}^\mathrm{tar}$ (``pasted image'' in Fig.~\ref{fig:overview}) may result in textural discontinuities at the boundaries of $\boldsymbol{M}^\mathrm{geo}$ in the final output.
To address this issue, our objective here is to derive a latent representation that effectively translates the geometry image into the visual domain of the target image through inversion.
This domain-shifted latent representation facilitates seamless blending within the latent space, thereby enabling a more naturalistic geometry transplantation.
It should be noted that since we leverage SD, the diffusion process operates in the latent space encoded by the variational autoencoder (VAE) instead of the pixel space.

To achieve this, we simultaneously invert the geometry and target images by Denoising Diffusion Implicit Models (DDIM) inversion~\cite{DDIM}, with replacing self-attention for the geometry image with a novel attention computation which we call Texture-aligning Attention, which additionally incorporates target image information.
Let $\boldsymbol{z}$ denote the latent representations in the VAE-encoded space, where specific instances will be distinguished by appropriate subscripts in the subsequent discussion.
The standard self-attention computation is formulated as:
\begin{equation}
  A(\boldsymbol{z}) = \softmax\left(\frac{\boldsymbol{Q}\boldsymbol{K}^\top}{\sqrt{d}}\right)\boldsymbol{V},
  \label{eq:self-attention}
\end{equation}
where $A(\cdot)$ denotes the functional representation of self-attention computation, where $d$ and $\top$ denote the dimension of the latent space and the transposition of a matrix, respectively, and where $\boldsymbol{Q}$, $\boldsymbol{K}$, and $\boldsymbol{V}$ represent the query, key, and value, respectively, all computed from the latents $\boldsymbol{z}$ corresponding to the same image.
Note that the diffusion step $t$ and the layer index $l$ are described in Fig.~\ref{fig:overview}, but omitted here for readability.
To align the source image domain with the target image domain, we replace the self-attention computation in the inversion of the geometry image with the following formula:
\begin{equation}
  A_\mathrm{TA}(\boldsymbol{z}^\mathrm{geo}; \boldsymbol{z}^\mathrm{tar}) = 
    \softmax
        \left(
            \frac
                {\boldsymbol{Q}^\mathrm{geo}}
                {\sqrt{d}}{\begin{bmatrix} \boldsymbol{K}^\mathrm{geo} \\ \boldsymbol{K}^\mathrm{tar} \end{bmatrix}}^\top
        \right)
        \begin{bmatrix} \boldsymbol{V}^\mathrm{geo} \\ \boldsymbol{V}^\mathrm{tar} \end{bmatrix},
  \label{eq:inversion-attention}
\end{equation}
where $A_\mathrm{TA}(\cdot)$ denotes Texture-aligning Attention, and where the superscripts $\mathrm{geo}$ and $\mathrm{tar}$ denote whether the respective $\boldsymbol{Q}$, $\boldsymbol{K}$, and $\boldsymbol{V}$ are computed from the geometry image feature map $\boldsymbol{z}_\mathrm{geo}$ or the target image feature map $\boldsymbol{z}_\mathrm{tar}$.
As previously elucidated, the geometry image preserves the relative color variations associated with surface geometry through the color adjustment, while simultaneously approximating the visual characteristics of the target image. 
Consequently, by retaining both the conventional self-attention key ($\boldsymbol{K}^\mathrm{geo}$) and value ($\boldsymbol{V}^\mathrm{geo}$) alongside their counterparts derived from the target image ($\boldsymbol{K}^\mathrm{tar}$ and $\boldsymbol{V}^\mathrm{tar}$), we anticipate a synergistic effect.
This dual-source attention mechanism is expected to efficiently facilitate both geometry preservation and material transformation during the inversion process.
A noisy latent representation $\boldsymbol{z}^\mathrm{geo}_T$ of the geometry image $\boldsymbol{I}^\mathrm{geo}$ is then 
obtained as a result of $T$ iterations of the inversion 
step using this custom attention.

Inversion of the source image $\boldsymbol{I}^\mathrm{src}$ and the target image $\boldsymbol{I}^\mathrm{tar}$ is performed independently, 
employing the standard self-attention mechanism without any modifications. 
The resulting noisy latent representations $\boldsymbol{z}^\mathrm{src}_T$ and $\boldsymbol{z}^\mathrm{tar}_T$ are then stored for subsequent use 
in the generation stage and in the blending stage, respectively. 
\raggedbottom
\paragraph{Blending.}
Utilizing the latent representations obtained in the inversion stage, 
in the blending stage (see the lower-middle panel in ~\cref{fig:overview}) 
we perform a latent-space blending between the geometry and target images.
Let $\boldsymbol{m}^\mathrm{geo}$ be the target mask $\boldsymbol{M}^\mathrm{geo}$ 
resized to the latent space.
The blended latent representation $\boldsymbol{z}_T^\mathrm{out}$ to be used for generating a synthesized image is computed from the latents $\boldsymbol{z}_T^\mathrm{geo}$ 
and $\boldsymbol{z}_T^\mathrm{tar}$ for the geometry and target images 
at diffusion step $T$ as follows:
\begin{equation}
  \boldsymbol{z}_T^\mathrm{out} = \boldsymbol{z}_T^\mathrm{geo} \odot \boldsymbol{m}^\mathrm{geo} + \boldsymbol{z}_T^\mathrm{tar} \odot \boldsymbol{m}^\mathrm{geo}_{\mathrm{comp}},
  \label{eq:latent-blend}
\end{equation} 
where $\boldsymbol{m}^\mathrm{geo}_{\mathrm{comp}}$ denotes the complementary 
mask of $\boldsymbol{m}^\mathrm{geo}$, defined by replacing 0 with 1 
and vice versa in $\boldsymbol{m}^\mathrm{geo}$. 

\paragraph{Generation.}
In the generation stage (see the lower-right panel in ~\cref{fig:overview}), 
we obtain the final output image by denoising the blended latent code 
$\boldsymbol{z}^\mathrm{out}_T$ synthesized in the blending stage.
Although we initialize the generation process with the latents that explicitly incorporate a portion of the source image, it turns out that applying standard denoising procedures may potentially result in the undesirable removal of the transplanted content along with the noise, leading to the loss of the geometry.

To address this challenge, we introduce a novel attention computation named Geometry-preserving Attention.
Specifically, we concurrently perform denoising on the source image latent representation $\boldsymbol{z}_T^\mathrm{src}$ obtained during the inversion stage and replace the self-attention computation for the blended latent representation $\boldsymbol{z}_T^\mathrm{out}$ with Geometry-preserving Attention by augmenting the key and value components with information derived from the source image. 
The process is formally expressed by the following equation:
\begin{equation}
  A_\mathrm{GP}(\boldsymbol{z}^\mathrm{geo}; \boldsymbol{z}^\mathrm{tar}) = 
    \softmax
        \left(
            \frac
                {\boldsymbol{Q}^\mathrm{geo}}
                {\sqrt{d}}{\begin{bmatrix} \boldsymbol{K}^\mathrm{out} \\ \hat{\boldsymbol{K}}^\mathrm{src} \end{bmatrix}}^\top
        \right)
        \begin{bmatrix} \boldsymbol{V}^\mathrm{out} \\ \hat{\boldsymbol{V}}^\mathrm{src} \end{bmatrix},
  \label{eq:gen-attention1}
\end{equation}
\begin{equation}
\hat{\boldsymbol{K}}^{\mathrm{src}} = \{K^{\prime\mathrm{src}}_{i} \mid m^{\mathrm{src}}_{i} \neq 0, i \in \{1, \ldots, H\times W\}\}, \label{eq:gen-attention3}
\end{equation}
\begin{equation}
      \hat{\boldsymbol{V}}^\mathrm{src} = \{V^{\prime\mathrm{src}}_{i} \mid m^{\mathrm{src}}_{i} \neq 0, i \in \{1, \ldots, H\times W\}\}.
  \label{eq:gen-attention4}
\end{equation}
Here, $A_\mathrm{GP}$ represents the functional form of the Geometry-preserving Attention.
The superscripts $\mathrm{out}$ and $\mathrm{src}$ denote that the corresponding variables are associated with the $\boldsymbol{z}^\mathrm{out}$ or $\boldsymbol{z}^\mathrm{src}$, respectively.
It is important to note that $K^{\prime\mathrm{src}}_{i}$ and $V^{\prime\mathrm{src}}_{i}$ are used instead of $K^\mathrm{src}_{i}$ and $V^\mathrm{src}_{i}$ to distinguish them from the keys and values used during the inversion process.
$H$ and $W$ denote the height and width of the latent feature map $\boldsymbol{z}^\mathrm{src}$, 
$\boldsymbol{m}^{\mathrm{src}} \in \{0,1\}^{H\times W}$ represents 
the source mask $\boldsymbol{M}^\mathrm{src}$ resized to the latent space, and  $\boldsymbol{m}^{\mathrm{src}}_i \in \{0, 1\}$ represents $i$-th element of it. 

\section{Experiments}
\label{sec:experiment}
\subsection{Setup}
We use the publicly available SD inpainting model checkpoint on HuggingFace\footnote{https://huggingface.co/runwayml/stable-diffusion-inpainting} as the backbone network.
For our experiments, both source image $\boldsymbol{I}^\mathrm{src}$ and target image $\boldsymbol{I}^\mathrm{tar}$ are cropped to a uniform size of $512\times512$ pixels.
When performing color adjustment as expressed in ~\cref{eq:color-adjustment-in-editing}, we set the scalar parameter $a$ to $0.5$.
We set the number of diffusion steps $T$ to $25$ and employ the DDIM sampler for both the inversion and generation processes, with no prompt input (i.e., an empty string is used) in either phase.

During the inversion phase, we leverage the insights from Garibi \etal~\cite{renoise}, which demonstrates that iterative inversion sampling at each diffusion step enhances inversion accuracy.
Specifically, we perform 5 sampling iterations per diffusion step in the inversion procedure.

\begin{figure*}[ht]
\centering
\includegraphics[width=1.0\linewidth]{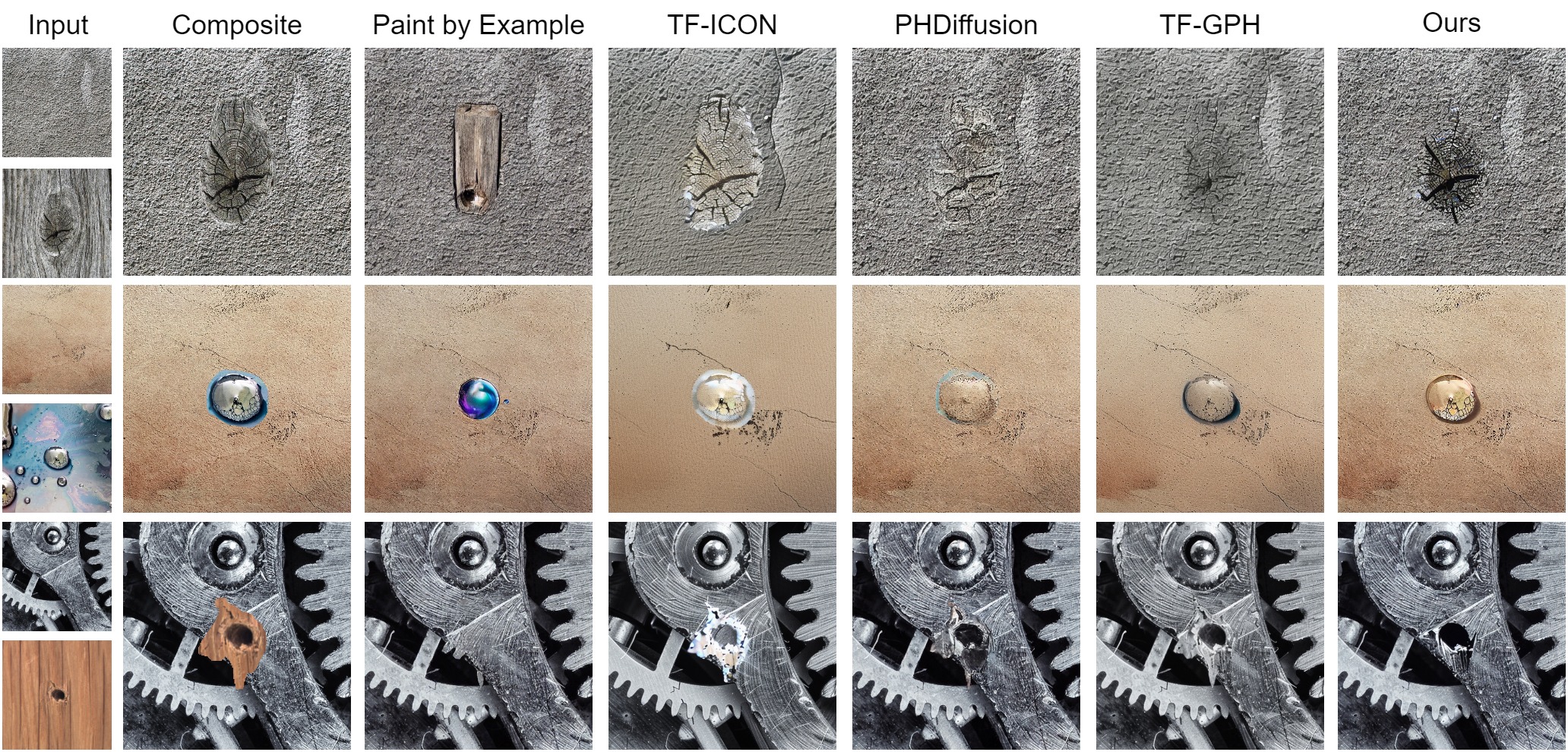}
\caption{
    Qualitative comparison of image generation results for geometry transfer. The methods to be compared are set to maximize the quality of the generated images, and the highest quality results are selected and included. Each column shows the output result by the compared method for harmonizing and the output result by our method.
    }
\label{fig:comparison}
\vspace{-0.4cm}
\end{figure*}

\subsection{Datasets and Metrics}
We use images from MVTec AD~\cite{mvtec1, mvtec2} and Pixabay\footnote{https://pixabay.com/} in our experiments. For the quantitative evaluation, we employ the following three metrics: LPIPS\cite{metric_LPIPS}, CLIP\cite{metric_CLIP}, and DISTS\cite{metric_DISTS}, since metrics specifically tailored for geometry transfer do not currently exist. These established metrics allow us to assess our method's performance in terms of image harmonization, and compare ours with existing methods. Each metric is evaluated against both the composite image background (denoted $\text{LPIPS}_\text{(bg)}$, $\text{CLIP}_\text{(bg)}$, and $\text{DISTS}_\text{(bg)}$) and the local region of the stretched and harmonized image (denoted $\text{LPIPS}_\text{(fg)}$, $\text{CLIP}_\text{(fg)}$, and $\text{DISTS}_\text{(fg)}$). Here, $\text{(bg)}$ and $\text{(fg)}$ refer to background and foreground, respectively. It is noteworthy that LPIPS and CLIP have previously been used as evaluation metrics in the existing literature\cite{TF-ICON, TF-GPH}. A total of 150 generated images are used for the quantitative performance evaluation, consisting of 10 composite foreground images and 15 background images, resulting in 150 combinations. Scores are calculated for each method, and the averages of the scores are presented.

\subsection{Baselines}
For comparison, we select four diffusion-model-based methods, including two image harmonization methods PHDiffusion~\cite{PHDiffusion} and TF-GPH~\cite{TF-GPH}, one image composition method TF-ICON~\cite{TF-ICON}, and one image editing method Paint by Example~\cite{PaintByExample}.
Since our Harmonizing Attention does not require prompts, we manually set suitable prompts for generating each sample with TF-ICON.
For all methods, we use the default or recommended hyperparameters.

\subsection{Qualitative Comparison}

We present a qualitative comparison of our proposed method, Harmonizing Attention, with four existing techniques: Paint by Example \cite{PaintByExample}, TF-ICON \cite{TF-ICON}, PHDiffusion \cite{PHDiffusion}, and TF-GPH \cite{TF-GPH}. Paint by Example and TF-ICON struggled with consistent texturing and maintaining original geometries. PHDiffusion showed better texture preservation but struggled with accurate geometry transfer. TF-GPH achieved a balance between texture and geometry alignment compared with previous methods, but still showed subtle texture inconsistencies and degradation of geometric properties upon closer inspection.

In contrast, our proposed method excelled in several critical aspects of image harmonization. It preserved the geometry and structural details of foreground objects with high fidelity. Simultaneously, it successfully adapted the texture and material properties to seamlessly match the background. This dual capability resulted in the most natural and realistic geometry integration across various scenarios, from textured surfaces such as wood to complex materials like metal parts.

\paragraph{User Study} To further compare the apparent quality of each technique, we conducted a user study with 105 participants from CrowdWorks\footnote{https://crowdworks.jp}. Participants rated 15 images, shown in ~\cref{fig:comparison}, on a scale from 1 to 5 (1 being the worst and 5 being the best) on the basis of the following criteria: background preservation (referred to as Quality Of Background, QOB), foreground preservation (referred to as Quality Of Foreground, QOF), and seamless composition (referred to as Quality Of Composition, QOC). Each qualitative rating score was calculated by applying weights to each response choice (5 for the best, 4, 3, 2, and 1 for the worst), and averaging all the responses.

The results are listed in ~\cref{tab:qual}. Our method consistently outperformed existing methods across all evaluated metrics.
The consistent superiority across these different metrics demonstrates the ability of our method to produce high-quality harmonized images with improved boundary definition, improved feature representation, and better overall content coherence compared with the existing techniques. 

\begin{table}[h]
	\caption{Qualitative evaluation results from user study. QOB, QOF, and QOC represent scores for background preservation, foreground preservation, and seamless composition, respectively. Higher scores indicate better performance across all criteria.}
        \vspace{-0.6cm}
	\begin{center}
        \begin{tabular}{>{\centering\arraybackslash\hspace{0pt}}p{3cm}|>{\centering\arraybackslash\hspace{0pt}}p{1.2cm}>{\centering\arraybackslash\hspace{0pt}}p{1.2cm}>{\centering\arraybackslash\hspace{0pt}}p{1.2cm}} \toprule
            Method & QOB $\uparrow$ & QOF $\uparrow$ & QOC $\uparrow$ \\ \midrule
            Paint by Example & 3.11 & 1.69 & 2.44 \\
            TF-ICON & 2.41 & 2.40 & 2.38 \\
            PHDiffusion & 3.10 & 2.64 & 3.26 \\
            TF-GPH & 3.12 & 2.79 & 3.40 \\
            Ours & \textbf{3.44} & \textbf{2.96} & \textbf{3.56} \\
            \bottomrule
		\end{tabular}
	\end{center}
        \vspace{-0.7cm} 
	\label{tab:qual}
\end{table}

\subsection{Quantitative Comparison}
We conducted a comprehensive quantitative evaluation to assess our method's performance against state-of-the-art approaches for geometry composition in target background images. ~\cref{tab:quan} presents results across multiple metrics, including $\text{LPIPS}$, $\text{CLIP}$, and $\text{DISTS}$, for both background and foreground components.
Our method achieved the lowest $\text{LPIPS}_\text{(bg)}$ score of 0.266 and $\text{DISTS}_\text{(bg)}$ score of 0.179, indicating superior preservation of background structural integrity. Furthermore, ours excelled in foreground semantic consistency, evidenced by the highest $\text{CLIP}_\text{(fg)}$ score of 91.352. While PHDiffusion and TF-GPH marginally outperformed ours on $\text{CLIP}_\text{(bg)}$ and $\text{LPIPS}_\text{(fg)}$ respectively, our method maintained competitive scores across all metrics.
This balanced performance, particularly our excellence in semantic preservation coupled with strong perceptual and structural similarity scores, suggests the effectiveness of our approach in achieving harmonious image composites. These results collectively demonstrate our method's advancement of the state-of-the-art in image harmonization across multiple dimensions of visual quality.

\begin{table*}[t]
	\caption{Quantitative evaluation results for geometry composition in a given target background image. Arrows next to each score indicate score interpretation: $\downarrow$ lower is better, $\uparrow$ higher is better. The minimum value for $\text{LPIPS}$ and $\text{DISTS}$ is 0, and the maximum value for $\text{CLIP}$ is 100. A total of 150 images were used for the evaluation.}
	\vspace{-0.6cm}
	\begin{center}
        \begin{tabular}{>{\centering\arraybackslash\hspace{0pt}}p{4cm}|>{\centering\arraybackslash\hspace{0pt}}p{1.8cm}>{\centering\arraybackslash\hspace{0pt}}p{1.8cm}>{\centering\arraybackslash\hspace{0pt}}p{1.8cm}>{\centering\arraybackslash\hspace{0pt}}p{1.8cm}>{\centering\arraybackslash\hspace{0pt}}p{1.7cm}>{\centering\arraybackslash\hspace{0pt}}p{1.7cm}} \toprule
            Method & $\text{LPIPS}_\text{(bg)}\downarrow$ & $\text{LPIPS}_\text{(fg)}\downarrow$
                & $\text{CLIP}_\text{(bg)}\uparrow$  & $\text{CLIP}_\text{(fg)}\uparrow$
                & $\text{DISTS}_\text{(bg)}\downarrow$ & $\text{DISTS}_\text{(fg)}\downarrow$ \\ \midrule
            Paint By Example \cite{PaintByExample} & 0.424 & 0.273 & 70.694 & 84.348 & 0.256 & 0.241 \\
            TF-ICON \cite{TF-ICON} & 0.434 & 0.392 & 69.470 & 81.422 & 0.309 & 0.315 \\ 
            PHDiffusion \cite{PHDiffusion} & 0.408 & 0.255 & \textbf{73.624} & 84.879 & 0.276 & \textbf{0.196} \\
            TF-GPH \cite{TF-GPH} & 0.324 & \textbf{0.252} & 71.234 & 86.936 & 0.223 & 0.237 \\ 
            Ours & \textbf{0.266} & 0.255 & 68.180 & \textbf{91.352} & \textbf{0.179} & 0.250 \\
            \bottomrule
		\end{tabular}
	\end{center}
	\vspace{-0.7cm} 
	\label{tab:quan}
\end{table*}

\subsection{Ablation Studies}
\subsubsection{Effectiveness of Color Shift}
We assessed our color shift method’s efficacy by varying the parameter $a$
in \cref{eq:color-adjustment-in-editing} and comparing it with histogram matching, a method that fully adjusts the source image’s color to match the target. The experiment was conducted under four conditions:
(i) no color adjustment (i.e., \eqref{eq:color-adjustment-in-editing} with $a=0.0$),
(ii) color adjustment via the proposed color shift method \eqref{eq:color-adjustment-in-editing} with $a=0.5$,
(iii) color adjustment via the proposed color shift method \eqref{eq:color-adjustment-in-editing} with $a=1.0$, and
(iv) color adjustment via histogram matching.
It should be noted that, except for color adjustment, all other settings were consistent across all three conditions.

\begin{figure}[htbp]
  \centering
   \includegraphics[width=1.0\linewidth]{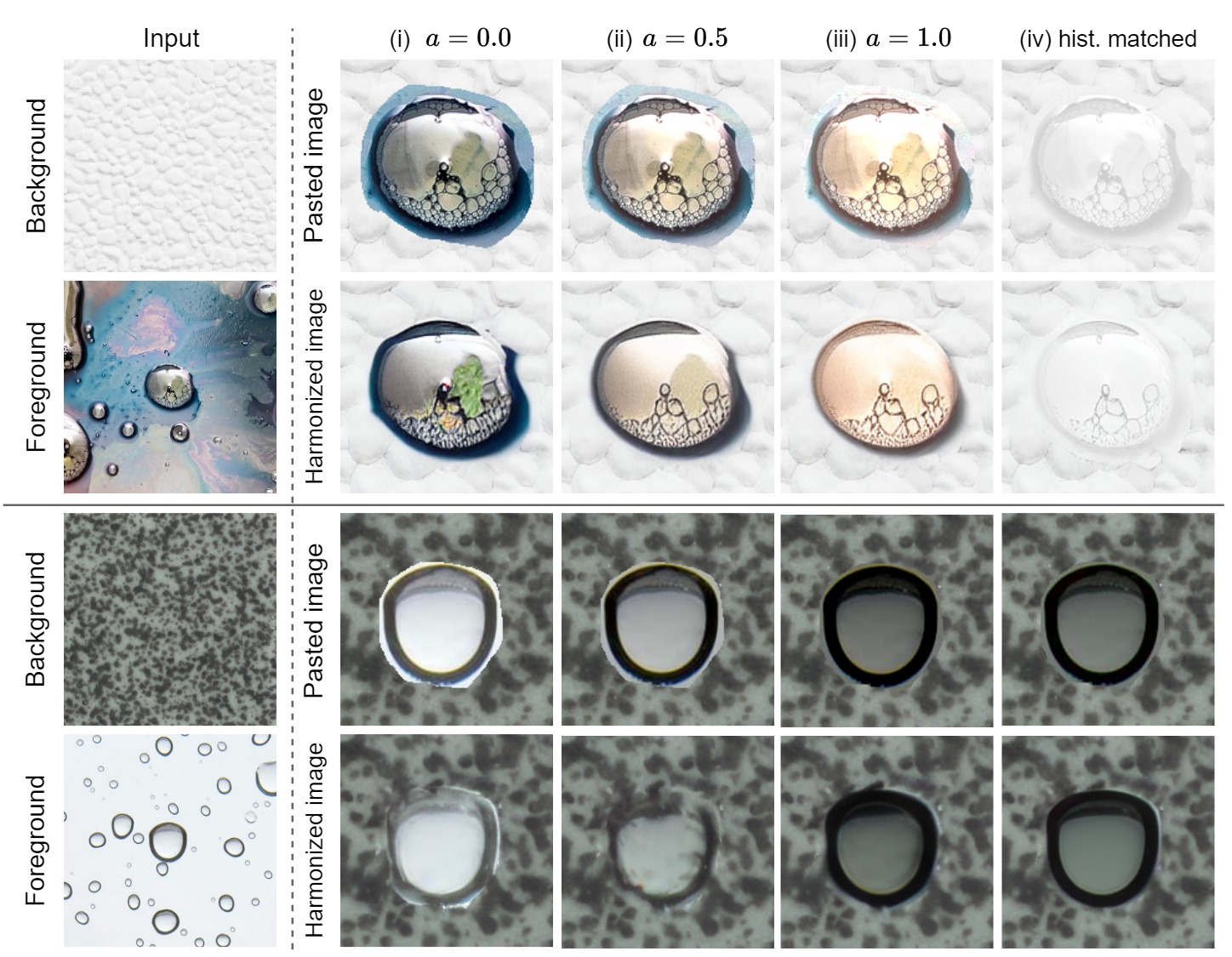}

   \caption{
   Qualitative comparison of images generated with different types of color adjustment. Here are examples of two paired images. The first columns are the background and foreground images to be input to our method. For each pair, the top line shows the pasted image with the foreground target attached to the background, and the bottom line shows the output from our method for each pasted image. Each line is the result of (i) using the original color image, (ii) and (iii) are color-shifted images of the target image using the color shift parameter, and (iv) using the color matched image with histogram matching method.
   }
   \label{fig:color-ablation}
   \vspace{-0.5cm}
\end{figure}

The qualitative results under these conditions are shown in ~\cref{fig:color-ablation}.
For reference, we also include pasted images, which are geometry images obtained under each condition directly pasted onto the target image.
Comparing (i) and (ii), (iii) in \cref{fig:color-ablation}, we can see that without color adjustment, the texture remained unchanged,
while with color shift, the geometry was reconstructed with a texture that was blended into the target image.
Also, comparing (ii), (iii) and (iv), we find that when histogram matching was used, the relative color changes within the source image area to be transplanted were not preserved,
resulting in loss of three-dimensionality.
Finally, comparing (ii) and (iii), a stronger color shift tends to improve texture blending, but tends to reduce the three-dimensionality of the object, making it difficult to maintain the original geometry.
These results suggest that a modest shift in the color of the source image to that of the target image is effective in transferring geometry.

\subsubsection{Effectiveness of Texture-aligning Attention}

We examined the effectiveness of Texture-aligning Attention.
As previously explained, the primary objective of the inversion phase is to align the geometry image, created from the source image, with the visual domain of the target image.
Intuitively, one might expect that using only the keys $\boldsymbol{K}^\mathrm{tar}$ and values $\boldsymbol{V}^\mathrm{tar}$ computed from the target image in ~\cref{eq:inversion-attention}, while eliminating $\boldsymbol{K}^\mathrm{geo}$ and $\boldsymbol{V}^\mathrm{geo}$ derived from the geometry image, would yield results more harmoniously integrated with the target image.
\begin{figure}[htbp]
  \centering
   \includegraphics[width=1.0\linewidth]{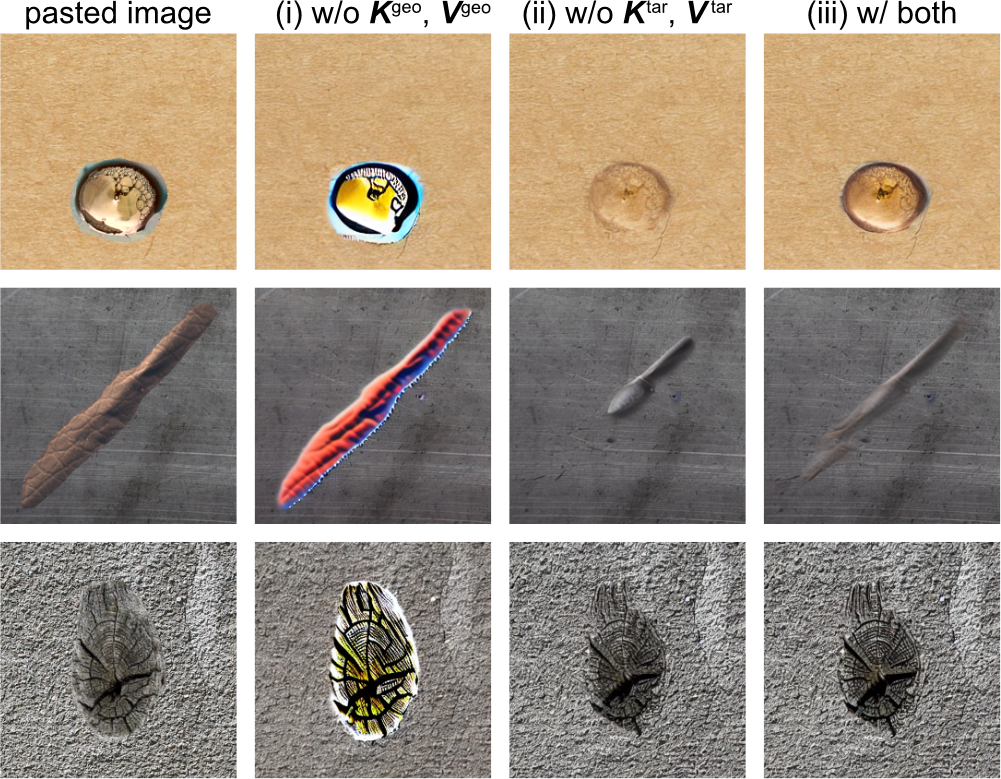}

   \caption{
   Qualitative comparison of images generated with different types of attention during the inversion phase.
   The first column shows pasted images, which we refer to as geometry images overlaid on target images, as shown in ~\cref{fig:overview}.
   The second column shows results for replacing $\boldsymbol{K}^\mathrm{geo}$ and $\boldsymbol{V}^\mathrm{geo}$ with $\boldsymbol{K}^\mathrm{tar}$ and $\boldsymbol{V}^\mathrm{tar}$.
   The third column shows results with the ordinary self-attention computation.
   The fourth column shows results with Texture-aligning Attention, where target- and geometry-derived keys and values are concatenated as shown in ~\cref{eq:inversion-attention}.
   }
   \label{fig:inversion-ablation}
   \vspace{-0.35cm}
\end{figure}
To evaluate the impact of each $\boldsymbol{K}$, $\boldsymbol{V}$ on this alignment, we conducted an ablation study comparing three inversion configurations:
(i) employing only target-image-derived attention (i.e., eliminating $\boldsymbol{K}^\mathrm{geo}$ and $\boldsymbol{V}^\mathrm{geo}$), and
(ii) utilizing only geometry-image-derived attention (i.e., removing $\boldsymbol{K}^\mathrm{tar}$ and $\boldsymbol{V}^\mathrm{tar}$).
(iii) using both geometry-image-derived and target-image-derived attention components,
It is noted that we used the same settings for editing, blending, and generation.

\Cref{fig:inversion-ablation} presents a comparative analysis of generation results under the three conditions.
Contrary to the initial intuition, the samples for condition (i) demonstrate that using only $\boldsymbol{K}^\mathrm{tar}$ and $\boldsymbol{V}^\mathrm{tar}$ does not lead to better integration with the target image.
Instead, as shown in condition (iii), using geometry-derived attention components yields results that are more harmoniously blended with the target image.
Even when comparing conditions with geometry-derived attention, the presence of target-derived attention tends to accentuate the source geometry.
A comparative analysis of columns (ii) and (iii) in \cref{fig:inversion-ablation} revealed notable distinctions.
In the first row, the absence of target-derived attention in column (ii) results in the disappearance of geometry.
Conversely, in the second row, column (ii) exhibits partial omission of geometry.
These observations suggest the effectiveness of employing both geometry-derived and target-derived attention in geometry transfer.

\subsubsection{Effectiveness of Geometry-preserving Attention}

\begin{figure}[htbp]
  \centering
   \includegraphics[width=1.0\linewidth]{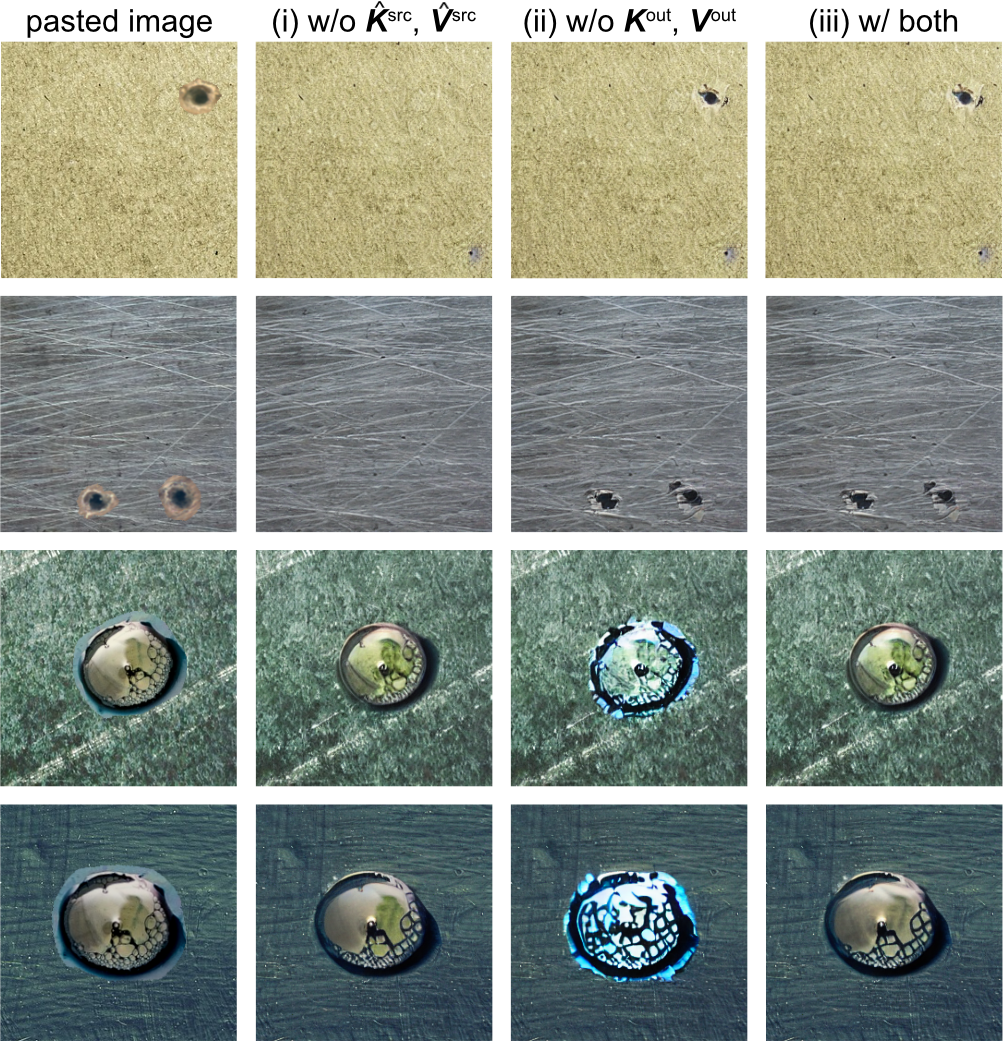}
   \caption{
    Qualitative comparison of images generated with different attention computation during the generation phase.
   The first column shows pasted images.
   The second column shows results for using only self-attention key $\boldsymbol{K}^\mathrm{out}$ and value $\boldsymbol{V}^\mathrm{out}$ during generation.
   The third column shows results for utilizing only source-image-derived key $\hat{\boldsymbol{K}}^\mathrm{src}$ and value $\hat{\boldsymbol{V}}^\mathrm{src}$.
   The fourth column shows results with Geometry-preserving Attention, where source- and self-derived keys and values are concatenated as shown in ~\cref{eq:gen-attention1}.
   }
   \label{fig:generation-ablation}
   \vspace{-0.35cm}
\end{figure}

We investigated the significance of Geometry-preserving Attention.
Specifically, we compared generation results under three conditions:
(i) employing only self-attention (i.e., eliminating $\hat{\boldsymbol{K}}^\mathrm{src}$ and $\hat{\boldsymbol{V}}^\mathrm{src}$), and
(ii) utilizing only source-image-derived attention (i.e., removing $\boldsymbol{K}^\mathrm{out}$ and $\boldsymbol{V}^\mathrm{out}$).
(iii) using both self-attention and source-image-derived attention components (i.e., using Geometry-preserving Attention as defined in ~\cref{eq:gen-attention1}).
\Cref{fig:generation-ablation} provides a qualitative analysis of the generation results.
Comparing (i) and (ii) or (iii), it was shown that the absence of source image attention leads to a significant degradation in geometry preservation, particularly evident in the transplantation of smaller regions.
As for the comparison of (ii) and (iii), samples in the third and fourth rows illustrate that only using source-image-derived attention results in overemphasized geometry when transplantation regions become larger.
These results indicate that although source-image-derived attention plays a key role in maintaining geometry information, the influence needs to be balanced by using self-attention simultaneously.

\subsection{Limitation}
The primary limitation of our work lies in its difficulty in transferring extremely large or small geometries. As evident from \cref{eq:inversion-attention,eq:gen-attention1}, the ratio of texture-related to geometry-related information in our customized attention calculation is strongly dependent on the size of the geometry transfer area. In addition, the generated output occasionally exhibits significant geometric deviations from intended results or produces unrelated content. This limitation might be attributed to our method's reliance on a pretrained SD model without additional training, potentially constraining its ability to fully capture texture and geometry.
Future improvements could focus on dynamically adjusting attention based on transfer area size, developing more robust attention mechanisms for diverse image combinations, and expanding generatable geometries through advanced training or architectural improvements. These refinements aim to address the current limitations and further improve the versatility and reliability of our approach across a wider range of geometry and image combinations.

\section{Conclusion}
\label{sec:conclusion}
In this work, we introduce Harmonizing Attention, a novel approach that facilitates the effective capture and transfer of material-independent geometry while preserving material-specific textural continuity. Our method uses custom Texture-aligning and Geometry-preserving Attention during inversion and generation processes, respectively, enabling the simultaneous referencing of source geometry and target texture information. Our approach achieves effective geometry transfer without requiring additional training or prompt engineering. Our method presented herein not only improves the creation of photorealistic composites but also expands the horizons of computer vision applications, ranging from augmented reality to advanced image editing.
{
\bibliographystyle{ieeetr}
\bibliography{egbib}
}

\end{document}